\documentclass[runningheads]{llncs}
\usepackage{graphicx}

\usepackage{tikz}
\usepackage{comment}
\usepackage{amsmath,amssymb} 
\usepackage{color}
\usepackage{hyperref}
\usepackage{url}
\usepackage{multirow}
\usepackage{booktabs}
\usepackage{makecell}
\usepackage{graphicx}
\usepackage{subfigure}
\usepackage{makecell}
\usepackage{marvosym}

\usepackage[accsupp]{axessibility}  

\newcommand{\orcid}[1]{\href{https://orcid.org/#1}{\includegraphics[width=10pt]{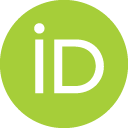}}}

\usepackage{hyphenat}

\begin{document}
\pagestyle{headings}
\mainmatter
\def\ECCVSubNumber{1757}  

\title{NDF: Neural Deformable Fields for Dynamic Human Modelling} 


\titlerunning{Neural Deformable Fields}
%
\author{Ruiqi Zhang\orcid{0000-0003-0095-3672} \and Jie Chen\orcid{0000-0001-8419-4620}\textsuperscript{\Letter}}
%
\authorrunning{Zhang and Chen}
%
\institute{Department of Computer Science, Hong Kong Baptist University \\
\email{\{csrqzhang, chenjie\}@comp.hkbu.edu.hk}}
\maketitle

\begin{abstract}
We propose Neural Deformable Fields (NDF), a new representation for dynamic human digitization from a multi-view video. Recent works proposed to represent a dynamic human body with shared canonical neural radiance fields which links to the observation space with deformation fields estimations. However, the learned canonical representation is static and the current design of the deformation fields is not able to represent large movements or detailed geometry changes. In this paper, we propose to learn a neural deformable field wrapped around a fitted parametric body model to represent the dynamic human. The NDF is spatially aligned by the underlying reference surface. A neural network is then learned to map pose to the dynamics of NDF. The proposed NDF representation can synthesize the digitized performer with novel views and novel poses with a detailed and reasonable dynamic appearance. Experiments show that our method significantly outperforms recent human synthesis methods. 

\end{abstract}

\section{Introduction}

Vision-based human performance capture has seen great progress in recent years due to fast development in both hardware and reconstruction algorithms like novel learning-based representation. It enables a wide variety of applications such as tele-presence, sportscast, and mixed reality. The enduring pandemic restricts our travel and public activities, which makes human performance digitization a research topic with great social and economic implications.

Human performance digitization can be roughly divided into human performance capture and human animation.
Traditionally, to achieve high-fidelity human performance capture including geometry and texture reconstruction, dense camera rigs~\cite{gortler1996lumigraph,hedman2018deep,joo2018total} and controlled lighting conditions~\cite{collet2015high,guo2019relightables} are required. These systems are extremely bulky and expensive, which limits their popularity. Nevertheless, these conventional capture systems could still fail under multi-person scenarios due to severe occlusion, which leads to ambiguity in appearance, pose, and motion sampling. After performance capture, human animation requires skilled artists to manually create a skeleton suitable for the human model and carefully design skinning weights~\cite{lewis2000pose} to achieve realistic animation, which requires countless human labor.

This paper aims to reduce the cost and improve the flexibility of human performance digitization. Many recent works have investigated the potential of neural implicit fields in novel view synthesis. NeRF~\cite{mildenhall2020nerf} proposed a neural implicit representation that can be effectively learned from multi-view images. The neural implicit representation is rendered to realistic images from novel views with volume rendering. However, NeRF has a high requirement for the camera numbers and it can only model a static scene which does not apply to multi-view videos of dynamic humans. To extend NeRF to dynamic scenes, an effective idea is to aggregate all observations over different video frames ~\cite{pumarola2021d,park2021nerfies,peng2021animatable,park2021hypernerf,li2021neural}. D-NeRF~\cite{pumarola2021d} and Nerfies~\cite{park2021nerfies} decompose a reconstruction into a canonical neural radiance field and a set of deformation fields that transform points in observation space to canonical space. To further simplify the learning of the deformation fields, Animatable NeRF~\cite{peng2021animatable} resorts to a parametric human body model as a strong geometry prior to the deformation fields. However, we claim that the current design of a shared canonical space and deformation fields prevents these methods from learning large movements and detailed geometry changes such as wrinkles of clothes as shown in the experiment.

To solve the above problems, rather than learning shared canonical neural radiance fields from multi-view videos, we use Neural Deformable Fields (NDF) to represent a dynamic scene. Specifically, we unwrap observation space to NDF space using the surface of a parametric body model as reference. NDF space is automatically aligned across frames and we further adopt the skeletal pose as posterior condition to model the dynamic changes. As a result, NDF space is more compact than the original observation space and it can model the dynamic changes caused by different poses. After training, we are able to animate the performer to different views and poses with a high degree of realism.

We evaluate our method on ZJU-MoCap~\cite{peng2021neural} and DynaCap~\cite{habermann2021real} datasets that capture dynamic humans in complex motions with synchronized cameras. The results show that our method can achieve high-fidelity reconstructions, especially for realistic dynamic changes in novel pose synthesis. The code is avaliable at \href{https://github.com/HKBU-VSComputing/2022_ECCV_NDF}{\url{https://github.com/HKBU-VSComputing/2022_ECCV_NDF}}.

In summary, the contributions of this paper are following:

\begin{itemize}
    \item We propose a compact novel representation called NDF, which can model the dynamic changes caused by different poses.
    \item The experiment results demonstrate significant improvement on the novel pose synthesis task, especially the detailed and realistic dynamic changes caused by different poses.
\end{itemize}

\section{Related Works}

\noindent\textbf{Learning-based Scene Representations}. According to the dimensionality of representation, several paradigms have been investigated for 3D content embedding in the context of image-based novel view synthesis. Multiplane image (MPI) \cite{zhou2018stereo,mildenhall2019local}, voxels \cite{sitzmann2019deepvoxels,lombardi2019neural}, point cloud \cite{aliev2020neural,dai2020neural}, and neural radiance fields \cite{mildenhall2020nerf,yu2021pixelnerf,martin2021nerf,garbin2021fastnerf,liu2020neuralvoxel} have all been under intense research focus recently. MPI learns scene representation in the form of fronto-parallel color and $\alpha$ planes, and novel views are rendered via homography-wraping. Sitzmann et al. \cite{sitzmann2019deepvoxels} proposed to learn a deepvoxel representation by dividing the 3D space into discrete 3D units that embed learned features, which was further replaced with a continuous learnable function \cite{sitzmann2019scene}. Mildenhall et al. \cite{mildenhall2020nerf} proposed to represent the scene as a neural radiance ﬁeld (NeRF) by directly mapping a continuous 5D coordinate to the volume density and view-dependent emitted radiance. NeRF has special advantages in that it can represent a continuous scene in arbitrary resolution and it can be effectively learned from multi-view images. Our method follows NeRF to reconstruct scenes from images and further extends it to dynamic scenes.

\noindent\textbf{Neural Implicit Representation for Human}. Habermann et al. \cite{habermann2021real} leverage a 3D scanned person-specific template to learn motion-dependent geometry as well as motion- and view-dependent dynamic textures from multi-view videos. The requirement of a high-quality 3D scanning restricted its use. Several recent works resort to learning a shared representation via deformable functions (in the form of NeRF \cite{pumarola2021d,park2021nerfies,shao2022doublefield,noguchi2021neural}). Restricted by the design choice of the function, it is difficult for these methods to model relatively large movements efficiently and they show limited generalizability to novel poses. Liu et al. \cite{liu2020neural} learns a person-speciﬁc embedding of the actor’s appearance given a monocular video and a textured mesh template of the actor. Neural Body \cite{peng2021neural} learns neural representations over the same set of latent codes anchored to the deformable human model SMPL \cite{loper2015smpl}, and naturally integrate observations across frames. The sparsity allows it to effectively aggregate observations across frames but the result shows it losses details like wrinkles of clothes. Neural Actor \cite{liu2021neural} learns an unposed implicit human model via inverse linear blend skinning functions (LBS). The model cannot handle surface dynamics and certain geometric information has been lost during the generation of 2D texture maps. Animatable NeRF~\cite{peng2021animatable} can animate the performer to novel poses however it requires fine-tuning on the novel pose frames. This would be impossible when applied to a completely novel pose that the performer has never done. Our method does not require fine-tuning and can be directly applied to completely novel poses after training.

\section{Proposed Method}

\begin{figure}
\centering
\includegraphics[width=0.95\linewidth]{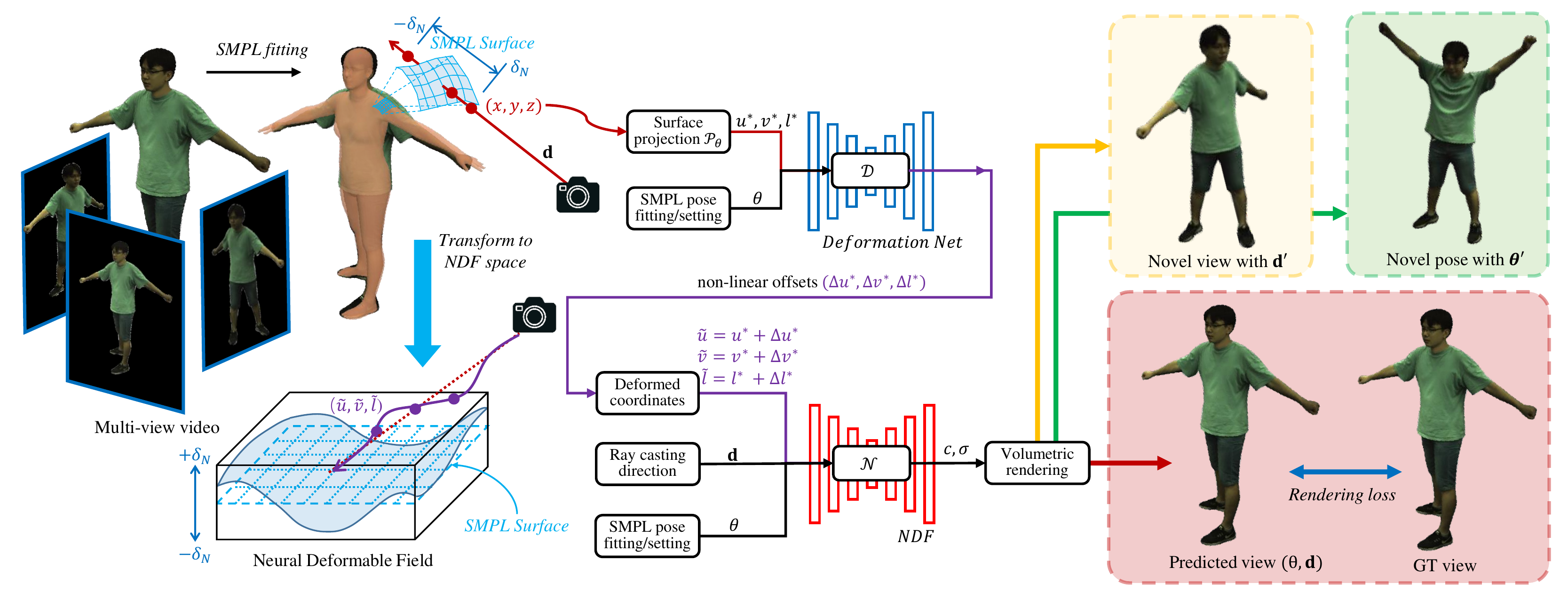}
\caption{\textbf{Overview of proposed method.} We query points in observation space, infer their densities and colors in NDF space and adopt volume rendering technique to synthesize images. For a given point $\mathbf{x}=(x,y,z)$ in observation space, we project it to NDF space with surface projection $\mathcal{P}_{\mathcal{\theta}}$ and further adopt deformation net $\mathcal{D}$ to slightly adjust the projection point $\mathbf{\tilde{u}}=(\tilde{u},\tilde{v},\tilde{l})$ in NDF space. A radiance field is then learned to predict the color $\mathbf{c}$ and density $\sigma$ for the point $\mathbf{\tilde{u}}$ in the unwrapped NDF space. The predicted color $\mathbf{c}$ and density $\sigma$ is then assigned back to the observation-space point $\mathbf{x}$. Finally, volume rendering is used to synthesize an image in the observation space.}
\label{fig:system}
\end{figure}

\textit{\textbf{Problem Setup.}} Given a training set of $T$-frame multi-view video of a dynamic human target over a sparse set of $K$ synchronized and calibrated cameras: $\mathcal{I}=\{I_t^k\}~(t=1\ldots T,k=1\ldots K)$, our goal is to digitize this performer using the proposed Neural Deformable Field (NDF) representation for both novel-view synthesis (NVS) and novel pose synthesis (NPS). 
Specifically, in the NVS task, we synthesize free-viewpoint renderings of the performance with novel camera angles. In the NPS task, we synthesize renderings with novel, unseen poses.

We build the NDF representation based on the state-of-the-art volumetric rendering model - Neural Radiance Field (NeRF) \cite{mildenhall2020nerf}, which predicts the color $\mathbf{c}$ and density $\sigma$ at spatial location $\mathbf{x}\in\mathbb{R}^3$ and view direction $\mathbf{d}\in\mathbb{S}^2$ via a neural network $\mathcal{F}$: $(\mathbf{x},\mathbf{d}) \mapsto(\mathbf{c},\sigma)$. Subsequently, volumetric rendering functions are used to render the final pixel color.
The differentiable rendering process enables optimization via comparing the output image with ground truth without 3D supervision. However, there are mainly two challenges in this setting. 
First, in our problem setup, only $K=4$ cameras are used, which is much less than what is sufficient to train a NeRF network. Second, due to the dynamic property of the human target, directly training a NeRF with all the frames will always cause artifacts and produce a coarse result.

To address these challenges, NDF fits a parametric human body model SMPL to associate 3D points among different video frames and learns a neural implicit field wrapped around and driven by the SMPL surface:
\begin{equation}
    \mathcal{N}: (\mathcal{D}(\mathcal{P}_{\mathcal{\theta}}(\mathbf{x})),\mathbf{d},\mathbf{\theta}) \mapsto(\mathbf{c},\sigma),
\end{equation}
where $\mathcal{P}_\theta$ is a projection function which projects a point's spatial location $\mathbf{x}$ to NDF space conditioned on the posed SMPL model with parameter $\theta$. $\mathcal{D}$ is a non-linear deformation function which keeps the surface continuity in the projection process. With the spatial alignment reference provided by the SMPL surface, NDF efficiently accumulates visual observations from the multi-view video frames; and given the strong geometry prior, NDF learns a geometry-guided field instead of a volume, which greatly reduces the learning complexity, leading to a much higher modelling efficiency. The details of each module will be introduced in this section.

\subsection{SMPL as Projection Reference with Non-linear Deformation}

To decrease NeRF's high requirement of camera numbers, a typical solution is to learn a deformation function $\Phi_t(\mathbf{x}):\mathbb{R}^3  \mapsto \mathbb{R}^3$ to map sample points $\mathbf{x}$ in frame $t$ to a shared canonical space~\cite{peng2021animatable}~\cite{pumarola2021d}. However, restricted by current design, these methods cannot deal with large movements or detailed geometry changes such as clothes wrinkles. To overcome these drawbacks, we resort to the texture map of SMPL as a reference to align 3D points across different frames and jointly train an integral NeRF model.

SMPL~\cite{loper2015smpl} is a skinned vertex-based model, which is defined as a function of shape parameters $\mathbf{\beta}$, pose parameters $\mathbf{\theta}$ and a rigid transformation $\mathbf{W}$ using Linear Blending Skinning (LBS). The template model $\mathbf{\Bar{T}}$ includes pre-defined 6890 vertices and their connections. With the pose-blend shape $B_P(\mathbf{\theta)}$ and shape-blend shape $B_S(\mathbf{\beta})$, the posed mesh $M(\mathbf{\theta},\mathbf{\beta})$ is got from the following equation:
\begin{align}
    M(\mathbf{\theta},\mathbf{\beta}) = \mathbf{W}(\mathbf{\Bar{T}}+B_S(\mathbf{\beta})+B_P(\mathbf{\theta})).
    \label{eq:smpl}
\end{align}
In this paper, we assume the posed mesh is pre-computed from the multi-view video and use the texture map of this mesh to conduct the projection function $\mathcal{P}_\theta$ from observation space to NDF space.

\subsubsection{Coordinates Projection.} \label{sec:projection}
As shown in Figure.~\ref{fig:system}, a 3D point $\mathbf{x}=(x,y,z)$ is projected to a point $\mathbf{u^*}=(u^*,v^*,l^*)$ in the \textit{unwrapped} Neural Deformable Fields (NDF) space with the projection function $\mathcal{P}_{\mathcal{\theta}}: \mathbf{x} \mapsto \mathbf{u}^*$. $\mathcal{P}_{\mathcal{\theta}}$ first projects the point $\mathbf{x}$ to the closest point $\mathbf{x'}\in\mathbb{R}^3$ on the fitted SMPL surface. $\mathbf{x'}$ has a 2D texel coordinate $(u^*,v^*)$ which is defined over SMPL's texture map and is calculated via:
\begin{align}
  (u^*,v^*,f^*) = \mathop{\arg\min}\limits_{u,v,f}\Vert \mathbf{x}-B_{u,v}(\mathcal{V}_{[\mathcal{F}(f)])}\Vert_2^2,
\end{align}
where $f \in \{1\ldots N_F\}$ is the triangle index, $\mathcal{V}_{[\mathcal{F}(f)]}$ is the three vertices of triangle $\mathcal{F}(f)$, $(u,v): u,v \in [0,1]$ are the texel coordinates on the texture map and $B_{u,v}(\cdot)$ is the barycentric interpolation function. SMPL is designed for modelling skinned human body and cannot capture surface dynamic changes. To model the dynamic geometry that deviates from the SMPL surface, we extend NDF to 3 dimensions with the euclidean distance $l^*$ between $\mathbf{x}$ and $\mathbf{x'}$ being the third dimension.

\subsubsection{Non-linear Deformation.} 
\begin{figure*}[t]
  \centering
  \includegraphics[width=.9\linewidth]{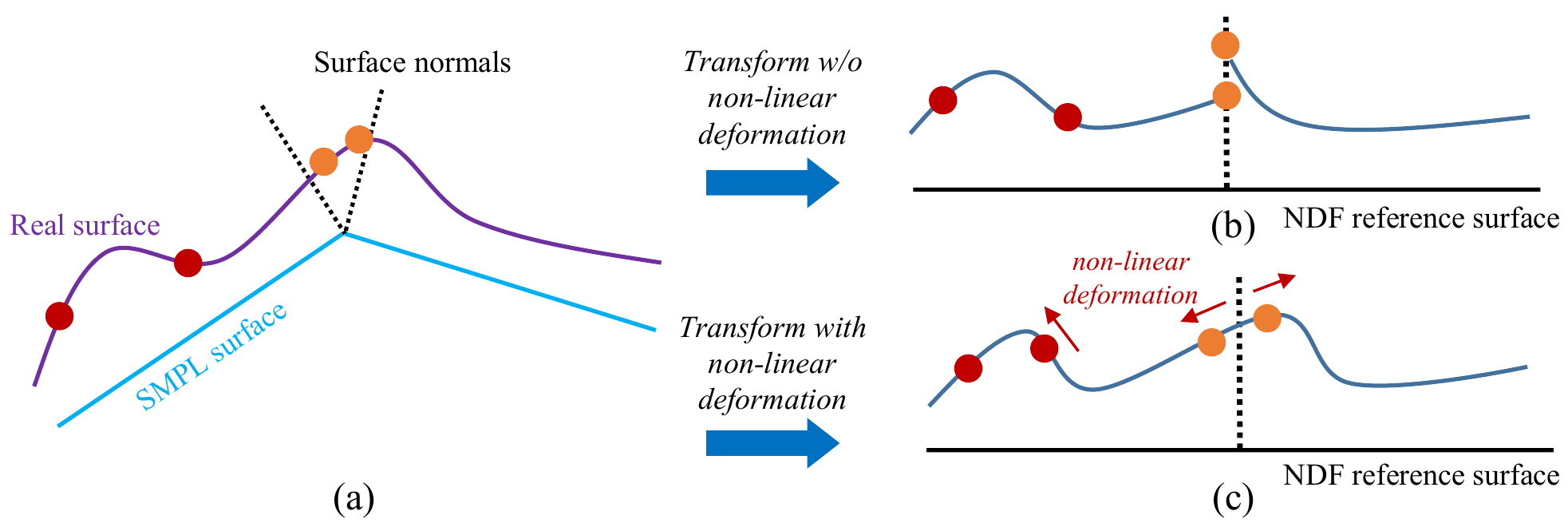}
\caption{A simplified 2D demonstration of transformation from the $(x,y,z)$ camera coordinates (a) to the $(u,v,l)$ NDF coordinates with and without non-linear deformation in (c) and (b), respectively.}
\label{fig:dnet}
\end{figure*}

We have projected an observation-space point $\mathbf{x}$ to $\mathbf{u^*}$ in NDF space using the UV coordinate of its nearest point on the SMPL surface as a reference. However, the continuous real surface will become discontinuous after projection. As shown in Figure~\ref{fig:dnet}(b), the two yellow points located on the continuous real surface in observation space will be closest to the same vertex on the SMPL surface if they locate in the same intersection of surface normals. After projection, the two yellow points will have the same $u^*,v^*$ but different $l^*$ in the NDF space. This will cause discontinuity at $(u^*,v^*)$ and hinder the learning of neural radiance fields. To solve this problem, we adopt a deformation net to slightly adjust the projection coordinate. As shown in Figure~\ref{fig:dnet}(c), this non-linear deformation can unwrap the surface fragment between the surface normal interval and the continuity of the real surface can be maintained. Formally, the deformed projection location $\mathbf{\tilde{u}}=(\tilde{u},\tilde{v},\tilde{l})$ is described as following:
\begin{gather}
    \triangle u^*, \triangle v^*, \triangle l^* = \mathcal{D}(\gamma_u(u^*,v^*,l^*),\theta) ,  \\
    \tilde{u},\tilde{v},\tilde{l} = u^*+\triangle u^*,v^*+\triangle v^*,l^*+\triangle l^* ,
\end{gather}
where $\mathcal{D}(\cdot)$ is the deformation net and $\gamma_u(\cdot)$ is the position embedding of $\mathbf{u^*}$. Note that the deformation aims to maintain the surface continuity in projection, but not to align points to a shared canonical space as in D-NeRF~\cite{pumarola2021d} and Nerfies~\cite{park2021nerfies}.

\subsection{Neural Deformable Fields} \label{sec:nidf}

\subsubsection{Rendering.} For a given 3D spatial location $\mathbf{x}$ along the target camera’s tracing ray 
direction $\mathbf{d}$, a point $\mathbf{\tilde{u}}$ will be found in the NDF space via projection and non-linear deformation as described above. The density for the point $\mathbf{x}$ will be estimated using an MLP $M_{\sigma}$: $\sigma(\mathbf{x})=M_{\sigma}(\gamma_u(\mathbf{\tilde{u}}),\theta)$. The color will be estimated with another MLP $M_c$: $c(\mathbf{x})=M_c(\gamma_u(\mathbf{\tilde{u}}), \gamma_d(\mathbf{d}),\theta)$, with an additional embedding $\gamma_d(\mathbf{d})$ for viewing direction, which ensures view-dependent effects.

The final image will be rendered via volumetric rendering \cite{kajiya1984ray} using numerical quadrature with $N$ consecutive samples $\{x_1,\ldots, x_N\}$ along the tracing ray:
\begin{equation}
I_{out} = \sum_{n=1}^{N}{(\prod_{m=1}^{n-1}{e^{-\sigma(\mathbf{x_m}) \cdot \delta_m}})\cdot (1-e^{-\sigma_n \cdot \delta_n}) \cdot c(\mathbf{x_n})}.
\label{eq:volumerender}
\end{equation}
Here $\delta_n=||\mathbf{x_n}-\mathbf{x_{n-1}}||_2$ denotes the quadrature segment along the ray.

\subsubsection{Geometry-guided Sampling Strategy.} To further facilitate the learning process of NDF, we use the fitted SMPL as geometry guidance to sample points more effectively and cancel the hierarchical sampling adopted in the original NeRF. Specifically, as shown in Figure~\ref{fig:system}, we take uniform samples but only accept samples if the projection distance $l^*$ is smaller than a hyper-parameter $\delta_N$.

\textbf{Remark.} NDF representation is lightweight, detailed, and intuitive. As compared with volumetric representations, its underlying geometrical linkage is well-defined by posed SMPL, resulting in reduced dimensionality for geometry reasoning, therefore significantly reducing model complexity and is much easier to train. The feature space of NDF span the whole UV dimension, which records much more details compared with Neural Body~\cite{peng2021neural}, where shared canonical features are only located at SMPL vertices.  By learning neural radiance fields conditioned on the pose, NDF can recover more intuitive dynamics related to changing pose rather than having to learn how to change query position in the canonical space through a per-frame deformation field like in Neural Actor~\cite{liu2021neural}.

\subsection{Deformable Fields for Novel Pose Synthesis} \label{sec:pose}

\subsubsection{Pose-driven NeRF}

By projecting points from the observation space to the NDF space, we are able to jointly learn a shared neural radiance field across frames. However, this representation would be only capable to capture a static geometry though it can be deformed to different poses. To model the dynamic change of human body geometry, we resort to the skeletal pose of SMPL as the posterior to infer the dynamic changes, i.e. we change the model from simply learning $\mathcal{N}: (\mathcal{D}(\mathcal{P}_{\mathcal{\theta}}(\mathbf{x})),\mathbf{d}) \mapsto(\mathbf{c},\sigma)$ to learning $\mathcal{N}: (\mathcal{D}(\mathcal{P}_{\mathcal{\theta}}(\mathbf{x})),\mathbf{d},\mathbf{\theta}) \mapsto(\mathbf{c},\sigma)$, where $\theta$ is the pose parameters of SMPL. In SMPL, the pose parameters $\theta$ is the axis-angle representation of the relative rotation of part $k$ with respect to its parent in the kinematic tree. Besides being used for changing pose, $\theta$ is also used to generate a pose-blend shape that describes the shape deformation caused by different poses. Inspired by this, we infer from pose $\theta$ the dynamics of the scene. In practice, we apply an additional feature extractor to extract high-level features of pose parameters which contain significantly more information than the pure pose parameters. The extracted pose features are then concatenated with the position embedding of $\mathbf{\tilde{u}}$ and fed into the following neural networks.

\subsubsection{Animation}\label{char:animation}

After training, NDF can be generalized to novel views or poses that do not occur in the training data $\mathcal{I}$. Specifically, given a viewing direction $\mathbf{d}$, a shape parameter $\mathbf{\beta}$ and a skeletal pose $\mathbf{\theta}$ got from a motion capture system or designed by hand, we calculate the mesh vertices through Equation~\ref{eq:smpl}. Then we sample points around the SMPL surface and render an image viewing from $\mathbf{d}$ with Equation~\ref{eq:volumerender}. 

\textbf{Remark.} NDF does not need to be fine-tuned on novel pose images compared with Animatable NeRF~\cite{peng2021neural} and can be applied to only sparse cameras compared with Neural Actor~\cite{liu2021neural}, where dense cameras are needed to pre-compute a realistic texture map. This animation ability only from sparse cameras would have a wide range of potential applications in VR or the metaverse.

\section{Experiment}

\subsection{Dataset and Metrics}

\textbf{ZJU-MoCap}~\cite{peng2021neural} records multi-view videos with 21 synchronous cameras and collects the shape parameters of SMPL as well as the global translation and the SMPL’s pose parameters with an off-the-shelf SMPL tracking system~\cite{EasyMocap_2021}. Following~\cite{peng2021neural}, we choose 9 sequences and 4 uniformly distributed cameras are used for training and the remaining cameras for testing. The video clips for evaluating novel view synthesis and novel pose synthesis are also the same with~\cite{peng2021neural}.

\noindent\textbf{DynaCap}. To further evaluate the generalization ability of our method, we select two sequences D1 and D2 from the DynaCap dataset~\cite{habermann2021real}. These two sequences record a performer with over 50 synchronous cameras. We fit neutral SMPL to these cameras using~\cite{EasyMocap_2021} and uniformly select 10 cameras for training and 5 cameras for testing.

\noindent\textbf{Metrics}. Following typical protocols~\cite{mildenhall2020nerf} and works most related to us~\cite{peng2021animatable}~\cite{peng2021neural}, we evaluate our method on image synthesis using two metrics: peak signal-to-noise ratio (PSNR) and structural similarity index (SSIM).

\subsection{Performance on NVS and NPS}

We compare our method with state-of-the-art view synthesis methods~\cite{peng2021neural,peng2021animatable} that also use SMPL models and can handle dynamic scenes. Neural Body~\cite{peng2021neural} represents the dynamic scene with an implicit field conditioned on a shared set of latent codes anchored on the vertices of SMPL and renders the images using volume rendering. Animatable NeRF~\cite{peng2021animatable} predicts the blend weights for each sample point and aggregates observations across frames to a shared canonical representation and further improves on novel pose synthesis by fine-tuning on novel pose images. All methods train a separate network for each scene. 

\begin{figure*}[t]
  \centering
  \includegraphics[width=0.96\linewidth]{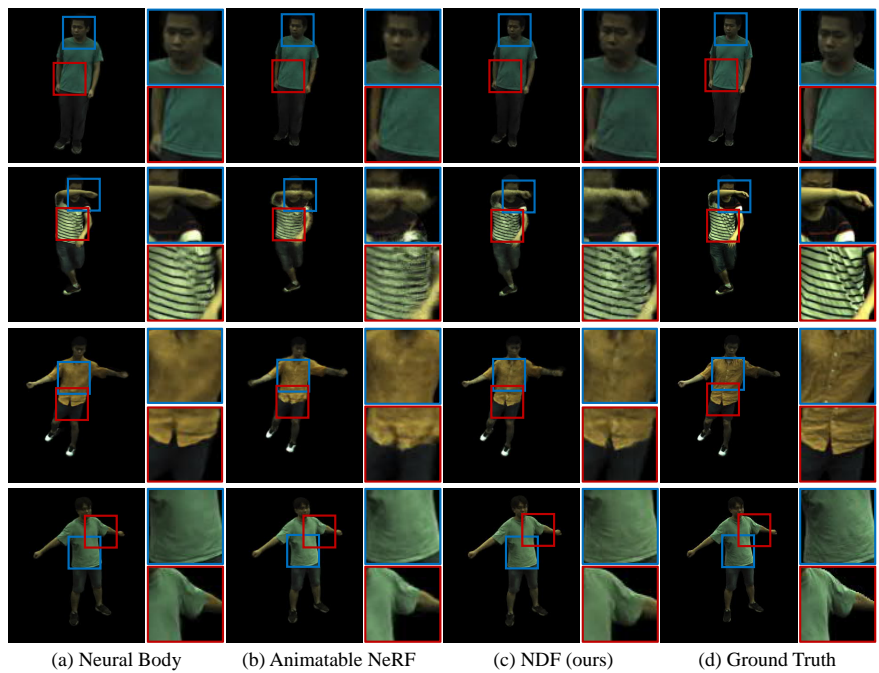}
\caption{Qualitative results of novel view synthesis on the ZJU-MoCap dataset.}
\label{fig:comparison_nv}
\end{figure*}

\begin{table}[ht]
  \begin{center}
    \caption{Results of novel view synthesis on the ZJU-MoCap dataset in terms of PSNR and SSIM (higher is better). “NB” means Neural Body. “AN” means Animatable NeRF. The best and the second best results are highlighted in red and blue, respectively.}
    \label{tab:novelView}
    \begin{tabular}{c|c|c|c|c|c|c}
        & \multicolumn{3}{c|}{PSNR} & \multicolumn{3}{c}{SSIM}\\
    \Xhline{1.2pt}
        & NB~\cite{peng2021neural} & AN~\cite{peng2021animatable} & OURS&NB~\cite{peng2021neural} & AN~\cite{peng2021animatable} & OURS \\
    \hline
      313 & {\color{red}30.39} & 29.27 & {\color{blue}29.84} & {\color{red}0.970} & 0.962 & {\color{blue}0.969}\\
      315 & {\color{red}26.53} & 24.22 & {\color{blue}25.71} & {\color{red}0.954} & 0.922 & {\color{blue}0.949}\\
      377 & {\color{red}27.49} & 26.63 & {\color{blue}26.85} & {\color{red}0.950} & 0.941 & {\color{blue}0.946}\\
      386 & {\color{red}28.66} & 26.78 & {\color{blue}28.21} & {\color{red}0.928} & 0.891 & {\color{blue}0.923}\\
      387 & {\color{red}25.52} & {\color{blue}24.75} & 24.52 & {\color{red}0.922} & {\color{blue}0.913} &0.911 \\
      390 & {\color{red}27.25} & 26.19 & {\color{blue}26.33} & {\color{red}0.920} & {\color{blue}0.915} &0.913 \\
      392 & {\color{red}29.41} & 27.79 & {\color{blue}28.40} & {\color{red}0.944} & 0.928 & {\color{blue}0.937}\\
      393 & {\color{red}27.41} & 26.06 & {\color{blue}26.73} & {\color{red}0.934} & 0.916 & {\color{blue}0.926}\\
      394 & {\color{red}28.65} & 27.53 & {\color{blue}27.98} & {\color{red}0.939} & 0.925 & {\color{blue}0.932}\\
    \hline
   average& {\color{red}27.92} & 26.58 & {\color{blue}27.17} & {\color{red}0.940} & 0.924 & {\color{blue}0.934}\\
    \end{tabular}
  \end{center}
\end{table}

\subsubsection{Evaluation on novel view synthesis.} Table~\ref{tab:novelView} shows the comparison of our method with Neural Body~\cite{peng2021neural} and Animatable NeRF~\cite{peng2021animatable} on ZJU-MoCap dataset. Our method outperforms Animatable NeRF~\cite{peng2021animatable} by a margin of 0.49 in terms of the PSNR metric and 0.01 in terms of the SSIM metric. It also performs close to Neural Body. Moreover, our method maintains its superiority when applied to DynaCap dataset as shown in Table~\ref{tab:D1_result}.

Figure~\ref{fig:comparison_nv} presents the qualitative comparison of our method with~\cite{peng2021neural,peng2021animatable} on the ZJU-MoCap dataset. Both~\cite{peng2021neural} and~\cite{peng2021animatable} have difficulty in recovering fine details of the dynamic scene. Neural Body~\cite{peng2021neural} turns to over-smooth the result as shown in the third person and the fourth person of Figure~\ref{fig:comparison_nv}. The clothes seam of the third person almost disappears and the small wrinkles on the clothes of the fourth person also disappear. Animatable NeRF~\cite{peng2021animatable} shows more artifacts as the blur of the first person's face and the second person's clothes. In contrast, our method can always recover realistic details like the hem of the third person.

\begin{figure*}[t]
  \centering
  \includegraphics[width=.96\linewidth]{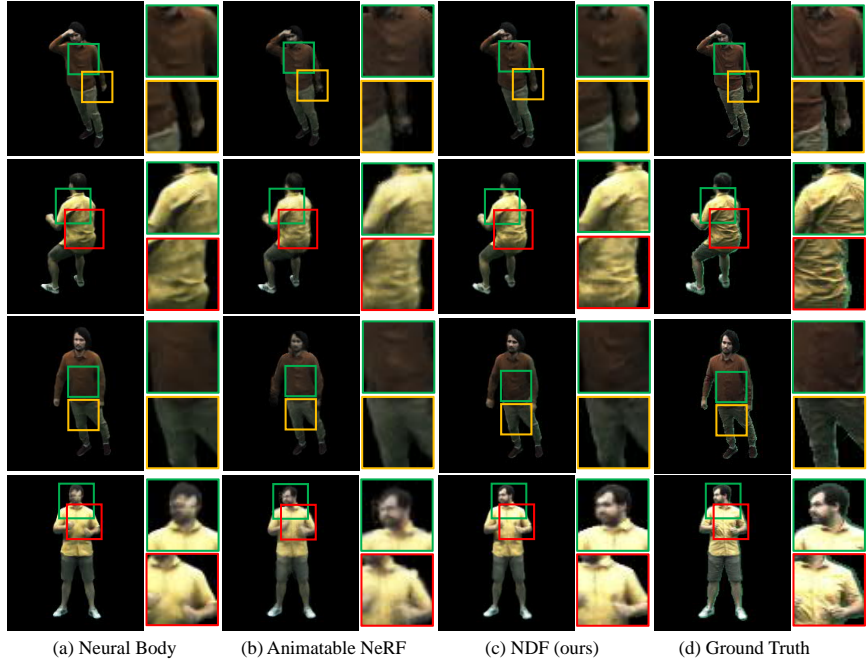}
\caption{Qualitative results of novel view synthesis and novel pose synthesis on the DynaCap dataset. Top 2 rows: novel view synthesis. Bottom 2 rows: novel pose synthesis.}
\label{fig:comparison_D1D2}
\end{figure*}

Figure~\ref{fig:comparison_D1D2} further presents the qualitative comparison on the DynaCap dataset. For the first two rows of novel view synthesis, our method can always recover realistic details. For the second row, Neural Body~\cite{peng2021neural} losses wrinkles on the back and Animatable NeRF~\cite{peng2021animatable} suffers from artifacts. While our method can reproduce high-quality wrinkles on the back.

\begin{figure*}[t]
  \centering
  \includegraphics[width=.96\linewidth]{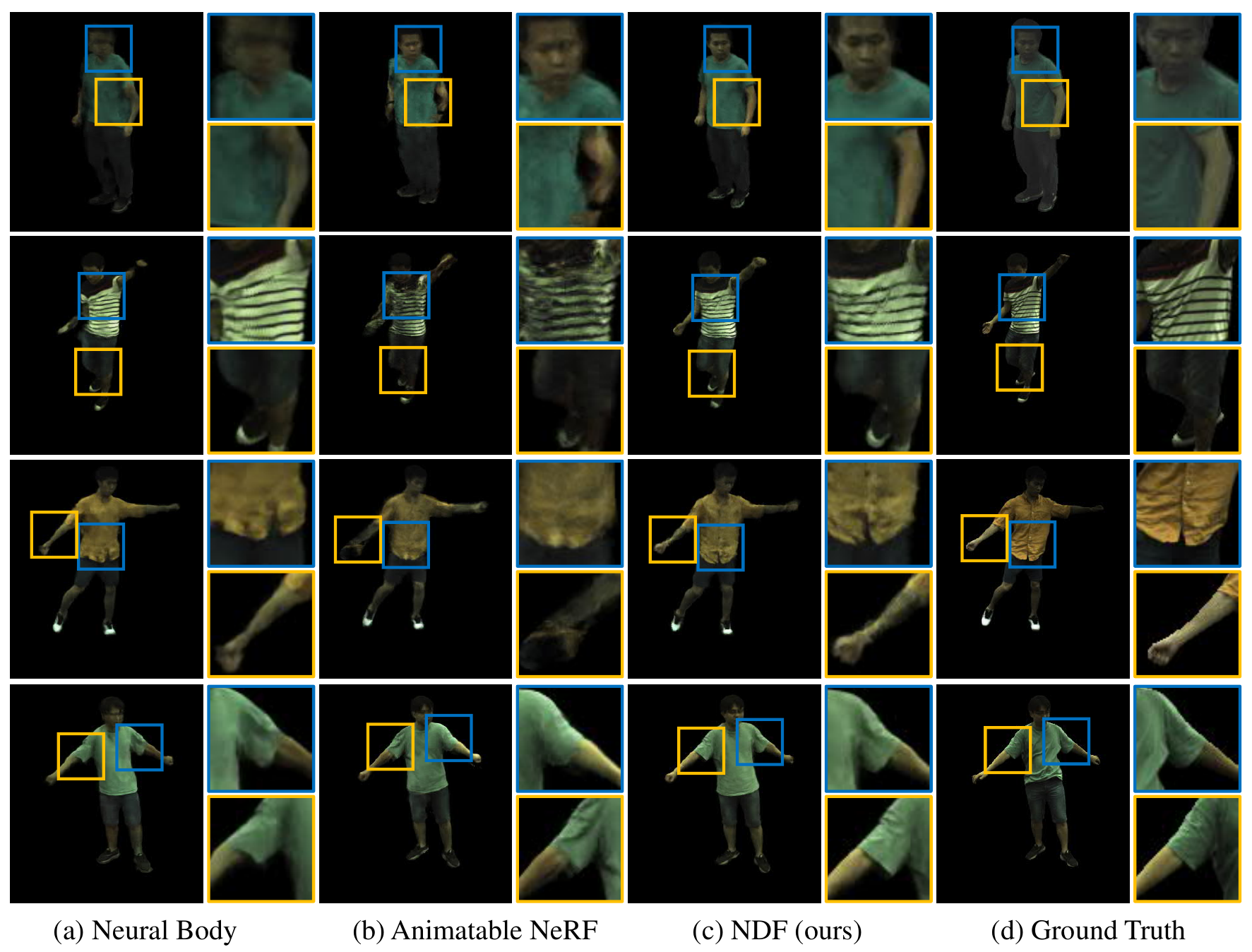}
\caption{Qualitative results of novel pose synthesis on the ZJU-MoCap dataset.}
\label{fig:comparison_np}
\end{figure*}

\begin{table}[ht]
  \begin{center}
    \caption{Results of novel pose synthesis on the ZJU-MoCap dataset in terms of PSNR and SSIM (higher is better)}
    \label{tab:novelPose}
    \begin{tabular}{c|c|c|c|c|c|c}
        & \multicolumn{3}{c|}{PSNR} & \multicolumn{3}{c}{SSIM}\\
    \Xhline{1.2pt}
        & NB~\cite{peng2021neural} & AN~\cite{peng2021animatable} & OURS&NB~\cite{peng2021neural} & AN~\cite{peng2021animatable} & OURS \\
    \hline
      313 & {\color{blue}23.49} & {\color{red}23.61} & 23.29 & 0.898 & {\color{red}0.908} & {\color{blue}0.903}\\
      315 & 19.38 & {\color{blue}19.45} & {\color{red}19.50} & 0.847 & {\color{blue}0.854} & {\color{red}0.857}\\
      377 & 23.89 & {\color{blue}25.03} & {\color{red}25.18} & 0.914 & {\color{blue}0.927} & {\color{red}0.928}\\
      386 & {\color{blue}25.63} & 25.14 & {\color{red}26.33} & 0.877 & {\color{blue}0.878} & {\color{red}0.893}\\
      387 & 21.75 & {\color{red}22.94} & {\color{blue}22.41} & 0.865 & {\color{red}0.892} & {\color{blue}0.880}\\
      390 & 23.81 & {\color{red}24.51} & {\color{blue}24.11} & 0.868 & {\color{red}0.889} & {\color{blue}0.881}\\
      392 & {\color{red}25.66} & 24.15 & {\color{blue}25.62} & {\color{blue}0.908} & 0.900 & {\color{red}0.914}\\
      393 & 23.30 & {\color{blue}23.97} & {\color{red}24.03} & 0.891 & {\color{blue}0.899} & {\color{red}0.902}\\
      394 & 23.76 & {\color{blue}24.29} & {\color{red}24.29} & 0.876 & {\color{red}0.893} & {\color{blue}0.890}\\
    \hline
   average& 23.41 & {\color{blue}23.68} & {\color{red}23.86} & 0.883 & {\color{blue}0.893} & {\color{red}0.894} \\
    \end{tabular}
  \end{center}
\end{table}

\begin{table}[ht]
  \begin{center}
    \caption{Results of novel view synthesis and novel pose synthesis on the DynaCap dataset in terms of PSNR and SSIM (higher is better).}
    \label{tab:D1_result}
    \begin{tabular}{c|c|c|c|c|c|c}
        & \multicolumn{3}{c|}{PSNR} & \multicolumn{3}{c}{SSIM}\\
    \Xhline{1.2pt}
        & NB~\cite{peng2021neural} & AN~\cite{peng2021animatable} & OURS&NB~\cite{peng2021neural} & AN~\cite{peng2021animatable} & OURS \\
    \hline
      novel view & {\color{blue}23.96} & 22.99 & {\color{red}24.73} & {\color{blue}0.889} & 0.872 & {\color{red}0.904}\\
      novel pose & {\color{blue}21.19} & 20.98 & {\color{red}21.42} & {\color{blue}0.828} & 0.828 & {\color{red}0.841}\\
    \end{tabular}
  \end{center}
\end{table}

\subsubsection{Evaluation on novel pose synthesis.} Table~\ref{tab:novelPose} shows the comparison of our method with Neural Body~\cite{peng2021neural} and Animatable NeRF~\cite{peng2021animatable} on novel pose synthesis. The result shows that our method outperforms compared method on most of the sequences and performs best for the average metrics. Note that Animatable NeRF~\cite{peng2021animatable} needs to be fine-tuned on novel pose images while our method can be directly applied to novel pose synthesis. 

The qualitative results are shown in Figure~\ref{fig:comparison_np}. Neural Body~\cite{peng2021neural} learns latent codes for training frames and does not model the dynamic change with respect to poses, thus it always suffers from artifacts when applied to novel pose synthesis. Though fine-tuned on novel pose images, Animatable NeRF~\cite{peng2021animatable} has difficulty in modelling large movements and also leads to blur result. Our method is able to recover details such as the hem of clothes for the third person even when applied to novel pose synthesis.

The bottom 2 rows of Figure~\ref{fig:comparison_D1D2} show the qualitative comparison on the DynaCap dataset. Neural Body~\cite{peng2021neural} fails to recover the face of the second person and Animatable NeRF produces severe artifacts on the face and hands, while our method can produce reliable realistic face and hands for the second person.

\begin{figure*}[t]
  \centering
  \includegraphics[width=0.9\linewidth]{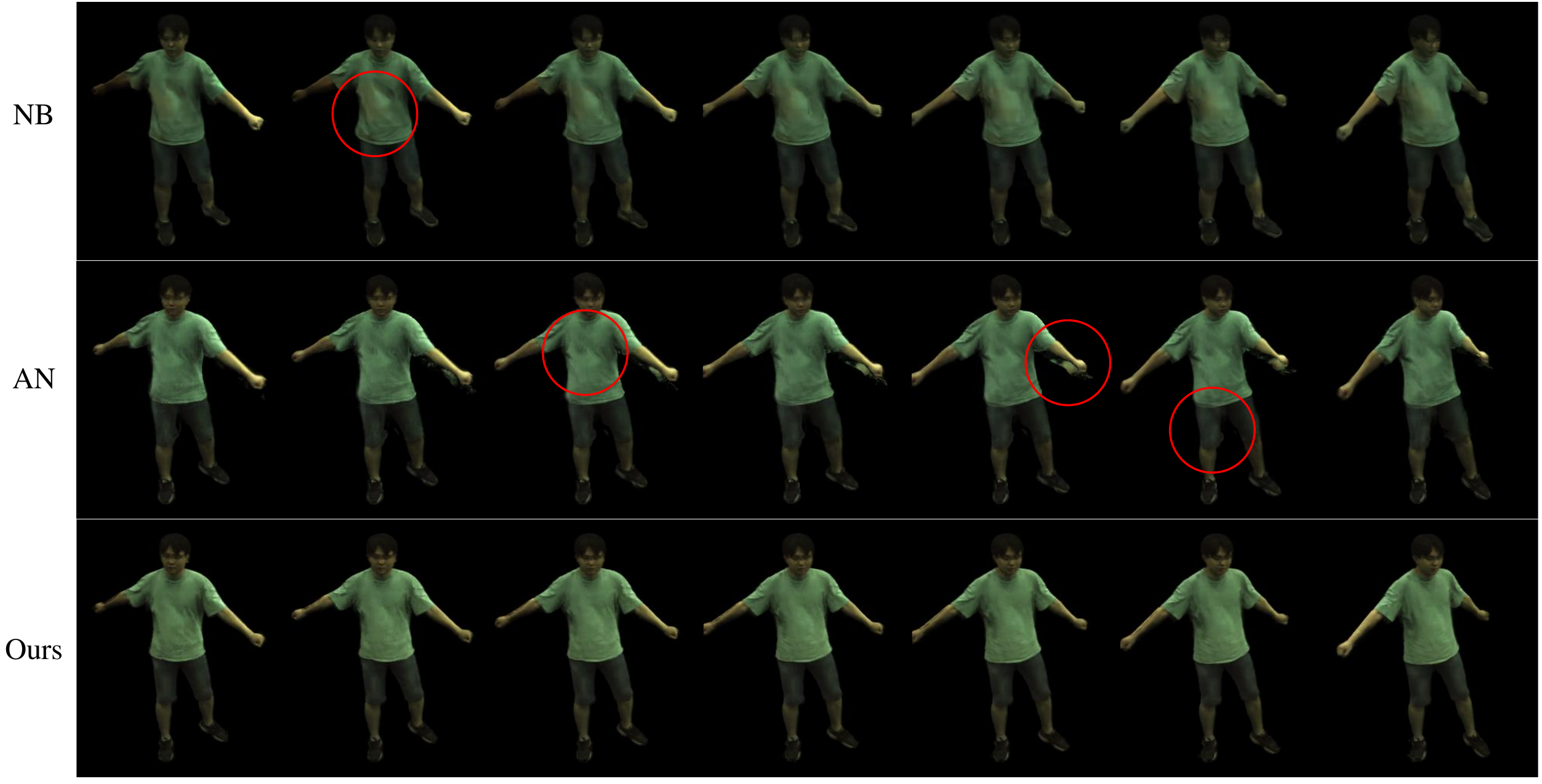}
\caption{Qualitative results of continuous frames to show temporal consistency. The red circles point out the flickering part of previous methods.}
\label{fig:temporal}
\end{figure*}

\subsection{Temporal Consistency}

NDF uses pose as condition which changes continuously and smoothly over time, while Neural Body and Animatable NeRF separately learn appearance codes for different frames. This endows NDF with better temporal consistency as can be seen from Figure~\ref{fig:temporal}. The red circles point out the flickering part of previous methods while NDF always shows better temporal consistency.

\begin{figure*}[t]
  \centering
  \includegraphics[width=1\linewidth]{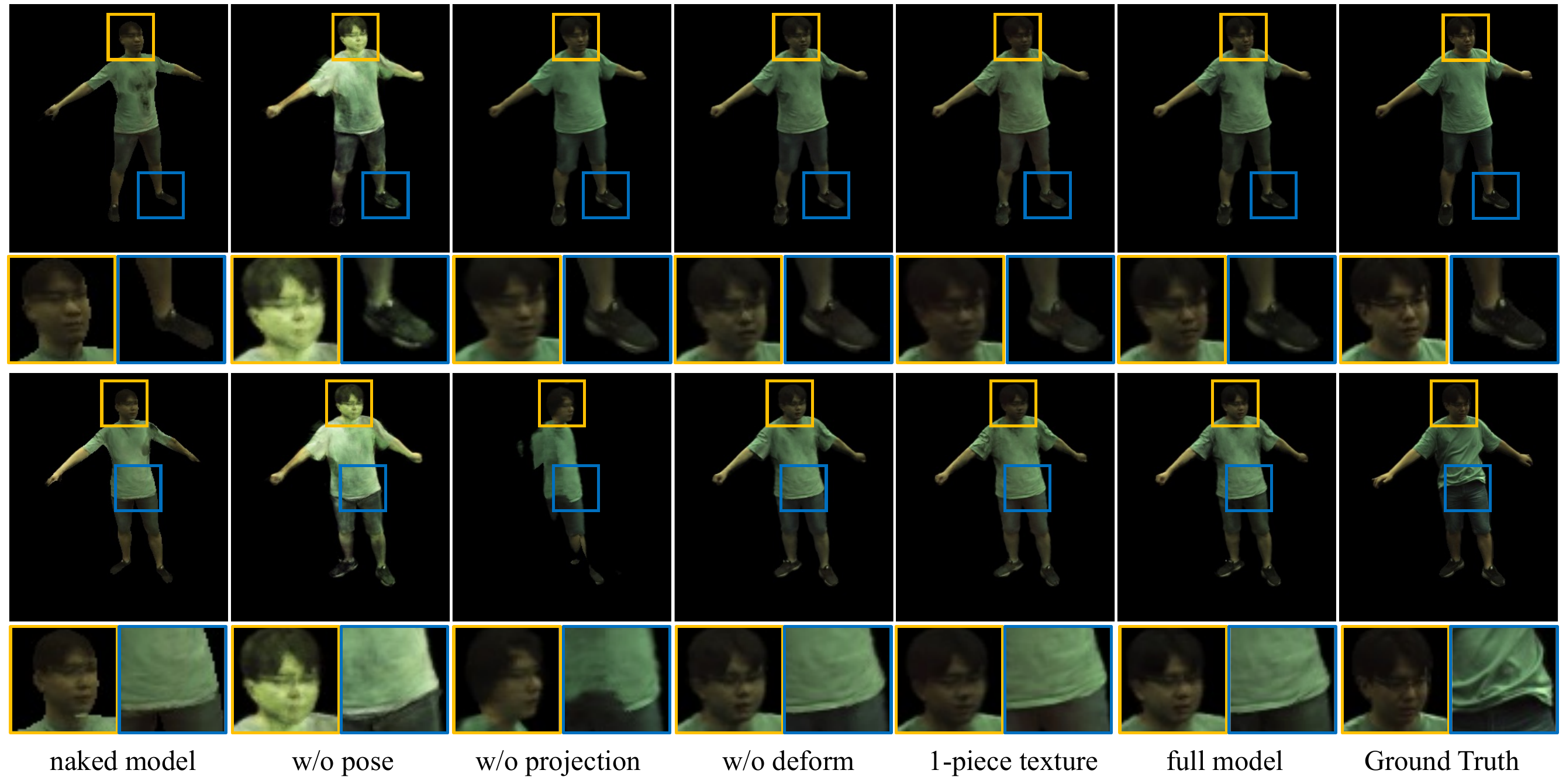}
\caption{Qualitative results of ablations. The first row and second row show the visual results for novel view synthesis and novel pose synthesis, respectively.}
\label{fig:ablation}
\end{figure*}

\begin{table}[ht]
  \begin{center}
    \caption{PSNR results of novel view synthesis and novel pose synthesis of ablations(higher is better).}
    \label{tab:ablation}
    \begin{tabular}{c|p{1.5cm}<{\centering}|p{1.5cm}<{\centering}|p{1.5cm}<{\centering}|p{1.5cm}<{\centering}|p{1.5cm}<{\centering}|p{1.5cm}<{\centering}}
        &\makecell[c]{naked\\model}&\makecell[c]{w/o\\pose}&\makecell[c]{w/o\\projection}&\makecell[c]{w/o\\deform}&\makecell[c]{1-piece\\texture}&\makecell[c]{full\\model}\\
    \Xhline{1.2pt}
      novel view & 23.65 & 21.98 & 29.73 & 29.72 & 29.93 & 29.75\\
      novel pose & 21.71 & 20.43 & 18.42 & 23.35 & 23.38 & 23.41\\
    \end{tabular}
  \end{center}
\end{table}

\subsection{Ablation Study}

We conduct ablation studies on one subject (313) of the ZJU-MoCap~\cite{peng2021neural} dataset in terms of the novel view synthesis and novel pose synthesis performance. We test the impact of the surface distance $\tilde{l}$, the impact of using pose as the condition to model dynamic change, the impact of projection from observation space to NDF space, the impact of deformation net, and the reliance of specific reference surface to show the effectiveness of our choice.

\subsubsection{Impact of the surface distance $\tilde{l}$ in NDF rendering.} To capture the dynamic geometry that cannot be captured by naked SMPL surface, we adopt the distance from a query point to its closest point on SMPL as the third dimension to model the NDF space as a field rather than a naked SMPL surface. To test the impact of this design, we only sample points on the SMPL surface thus the $\tilde{l}$ for projected points are all 0. As shown in the first column of Figure~\ref{fig:ablation} and Table~\ref{tab:ablation}, modelling the NDF space as naked SMPL surface causes severe artifacts, especially for clothes that cannot be captured by SMPL surface.

\subsubsection{Using pose as condition to model dynamic change.} In this experiment, we cancel using pose as the condition and jointly learn a shared canonical NDF for all frames. As shown in the second column of Figure~\ref{fig:ablation}, the model cannot handle dynamic changes and produces blur rendering at dynamic regions.

\subsubsection{Impact of projection from observation space to NDF space.} In this experiment, we directly use the observation-space coordinates $(x,y,z)$ as input to the neural network. The model needs to learn the mapping from pose to the whole 3D volume however it is severely difficult. As shown in the third column of Figure~\ref{fig:ablation}, though the model can synthesize novel views of the performer, it totally fails on novel pose synthesis.

\subsubsection{Impact of deformation net.} The deformation net aims to maintain the surface continuity after projection as claimed in Figure~\ref{fig:dnet}. As shown in the fourth column of Figure~\ref{fig:ablation}, the face and shoes become slightly noisier and we infer this is because the triangle surfaces of SMPL are small and dense on the face and feet. The result confirms the effectiveness of our design of the deformation net.

\subsubsection{Reliance of specific reference surface.} NDF does not rely on a specific texture map as the reference surface. To validate this, we replace the default texture map of SMPL with a self-designed texture map which can be found in the supplementary material. We cut the seam of the SMPL mesh in Blender and unwrap the mesh into one piece in the UV space. As shown in the fifth column of Figure~\ref{fig:ablation} and Table~\ref{tab:ablation}, with the 1-piece texture map as reference surface, the face becomes slightly blurred but the whole effect is still robust. This is because the UV region corresponding to the face occurs to be much smaller than in the default texture map of SMPL. The result shows that our method does not rely on a specific texture map and a self-designed texture map can also be used to unwrap points from observation space to NDF space.

\section{Limitations and Future Works}

Learning neural radiance fields conditioned on pose in NDF space enables us to obtain impressive performances on human digitization. However, our method has a few limitations. 1) Currently our method has a high requirement for the fitting effect of SMPL. Hopefully, in the future, we can integrate the fitting of SMPL in the pipeline and make the fitting and rendering benefit from each other. 2) In more complex scenes, the dynamic content depends both on pose and temporal information. A potential solution is to train the model with an auto-regressive way to model the relationship to temporal information.

\section{Conclusions}

We propose a novel representation of Neural Deformable Fields (NDF) to model dynamic humans. We unwrap observation space to NDF space using a parametric body model as a reference. Then a neural radiance field conditioned on skeletal pose is learned and volume rendering is used to render the pixel color. After training from multi-view videos, our method can synthesize the performer with arbitrary view direction and pose. Extensive experiments on ZJU-MoCap and DynaCap demonstrated that our method outperforms the state-of-the-art in terms of rendering quality and produces faithful pose-dependent appearance changes and wrinkle patterns.

\textbf{Acknowledgments}
The research was supported by the Theme-based Research Scheme, Research Grants Council of Hong Kong (T45-205/21-N).

\clearpage
%
%
\bibliographystyle{splncs04}
\bibliography{mybib}
\end{document}